\documentclass[12pt]{article}
\usepackage{fullpage}
\usepackage{graphicx}
\usepackage{psfrag}
\usepackage{amsmath}
\usepackage{amsfonts}
\usepackage{hyperref}
\usepackage{url}
\usepackage{booktabs}
\usepackage[bf]{caption}
\usepackage{subcaption}

\newcommand{\reals}{{\mbox{\bf R}}}

\newcommand{\eg}{{\it e.g.}}
\newcommand{\ie}{{\it i.e.}}

\newcommand{\dt}{{\hspace{1pt} dt}}
\newcommand{\Rcum}{R^\mathrm{cum}}
\newcommand{\Lamcum}{\lambda ^\mathrm{cum}}
\newcommand{\calRcum}{\mathcal R^\mathrm{cum}}
\newcommand{\Rinst}{R^\mathrm{inst}}
\newcommand{\calRinst}{\mathcal R^\mathrm{inst}}
\newcommand{\Laminst}{\lambda^\mathrm{inst}}
\newcommand{\Rinstso}{R^\mathrm{inst^2}}
\newcommand{\calRinstso}{\mathcal R^\mathrm{inst^2}}
\newcommand{\Laminstso}{\lambda^\mathrm{inst^2}}
\newcommand{\Smin}{s^\mathrm{min}}
\newcommand{\Smax}{s^\mathrm{max}}

\newcommand{\Ttrain}{t^{\mathrm{train}}}
\newcommand{\Ttest}{t^{\mathrm{test}}}

\newcommand{\Phitrue}{\phi^\mathrm{true}}

\newcommand{\gdtw}{{\href{https://github.com/dderiso/gdtw}{\texttt{GDTW}}}}

\newcommand{\BEAS}{\begin{eqnarray*}}
\newcommand{\EEAS}{\end{eqnarray*}}
\newcommand{\BEQ}{\begin{equation}}
\newcommand{\EEQ}{\end{equation}}
\newcommand{\BIT}{\begin{itemize}}
\newcommand{\EIT}{\end{itemize}}

\title{A General Optimization Framework for Dynamic Time Warping} 
\author{Dave Deriso \and Stephen Boyd}

\begin{document}
\maketitle

\begin{abstract}
The goal of \emph{dynamic time warping} is to transform or warp time
in order to approximately align two signals together. 
We pose the choice of warping function as an optimization problem with
several terms in the objective.
The first term measures the misalignment of the time-warped signals.
Two additional regularization terms penalize the cumulative warping and 
the instantaneous rate of time warping;
constraints on the warping can be imposed by assigning the value
$+\infty$ to the regularization terms.
Different choices of the three objective terms yield different time
warping functions that trade off signal fit or alignment and properties
of the warping function.
The optimization problem we formulate is a classical optimal control problem, 
with initial and terminal constraints, and a state dimension of one.  
We describe an effective general method that minimizes the objective by
discretizing the values of the original and warped time, and using
standard dynamic programming to compute the (globally) optimal warping
function with the discretized values. 
Iterated refinement of this scheme yields a high accuracy warping
function in just a few iterations.
Our method is implemented as an open source Python package \gdtw{}.
\end{abstract}

\section{Background}
The goal of dynamic time warping (DTW) is to find a time warping function
that transforms, or warps, time in order to approximately align two signals 
together~\cite{sakoe1978dynamic}.
At the same time, we prefer that the time warping be as gentle as
possible, in some sense, or we require that it satisfy some requirements.

DTW is a versatile tool used in many scientific fields, including 
biology, economics, signal processing, finance, and robotics.
It can be used to measure a realistic distance between 
two signals, usually by taking the distance between
them after one is time-warped. 
In another case, the distance can be the minimum amount of 
warping needed to align one signal to the other with some level of 
fidelity.
Time warping can be used to develop a simple model of a signal, or to 
improve a predictor; as a simple example, a suitable time warping can
lead to a signal being well fit by an auto-regressive or other model.
It can be employed in any machine-learning application that relies on 
signals, such as PCA, clustering, regression, logistic regression, or
multi-class classification. 
(We return to this topic in \S\ref{s-alignment}.)

Almost all DTW methods are based on the original DTW 
algorithm~\cite{sakoe1978dynamic}, which uses dynamic programming to 
compute a time warping path that minimizes misalignments in the 
time-warped signals while satisfying monotonicity, boundary, 
and continuity constraints.
The monotonicity constraint ensures that the path represents a monotone 
increasing function of time.
The boundary constraint enforces that the warping path beings with the 
origin point of both signals and ends with their terminal points. 
The continuity constraint restricts transitions in the path to adjacent 
points in time.

Despite its popularity, DTW has a longstanding problem with producing 
sharp irregularities in the time warp function that cause 
many time points of one signal to be erroneously mapped onto a single 
point, or ``singularity,'' in the other signal.
Most of the literature on reducing the occurrence of singularities falls 
into two camps: preprocessing the input signals, and variations on 
continuity constraints.
Preprocessing techniques rely on transformations of the input signals, 
which make them smoother or emphasize features or landmarks, to 
indirectly influence the smoothness of the warping function.
Notable approaches use combinations of first and second derivatives
\cite{keogh2001derivative,marron2015functional,singh2008optimization},
square-root velocity functions \cite{srivastava2011registration},
adaptive down-sampling \cite{dupont2015coarse},
and ensembles of features including wavelet transforms, derivatives, and 
several others \cite{zhao2016shapedtw}.
Variations of the continuity constraints relax the restriction on 
transitions in the path, which allows smoother warping paths to be chosen. 
Instead of only restricting transitions to one of three neighboring points 
in time, as in the original DTW algorithm, these variations expand the 
set of allowable points to those specified by a ``step pattern,'' of 
which there are many, including symmetric or asymmetric, types 
\emph{I-IV}, and sub-types \emph{a-d} 
\cite{sakoe1978dynamic,itakura1975minimum,myers1980performance,
rabiner1993fundamentals}.
While preprocessing and step patterns may result in smoother warping 
functions, they are \emph{ad-hoc} techniques that often require 
hand-selection for different types of input signals.

We propose to handle these issues entirely within an optimization framework
in continuous time. 
Here we pose DTW as an optimization problem with several penalty terms 
in the objective.
The basic term in our objective penalizes misalignments in the 
time-warped signals, while two additional terms penalize (and constrain) 
the time warping function. 
One of these terms penalizes the cumulative warping, which limits 
over-fitting similar to ``ridge'' or ``lasso'' regularization 
\cite{TA77,tibshirani1996regression}.
The other term penalizes the instantaneous rate of time warping, which
produces smoother warping functions, an idea that previously proposed in 
\cite{green1993nonparametric,ramsay1998functional,ramsay2007applied,
srivastava2016functional}.

Our formulation offers almost complete freedom in choosing the functions 
used to compare the sequences, and to penalize the warping function.
We include constraints on the fit and warping functions by allowing
these functions to take on the value $+\infty$.
Traditional penalty functions include the square or absolute value.
Less traditional but useful ones include for example the fraction of
time the two signals are within some threshold distance, or a minimum or
maximum on the cumulative warping function.
The choice of these functions, and how much they are scaled with respect
to each other, gives a very wide range of choices for potential time
warpings.

Our continuous time formulation allows for non-uniformly sampled signals,
which allows us to use simple out-of-sample validation techniques to help 
guide the choice of time warping penalties; 
in particular, we can determine whether a time warp is `over-fit'.
Our handling of missing data in the input signals is useful in itself 
since real-world data often have missing entries.
To the best of our knowledge, we are the first use of out-of-sample 
validation for performing model selection in DTW.

We develop a single, efficient algorithm that solves our formulation,
independent of the particular choices of the penalty functions. 
Our algorithm uses dynamic programming to exactly solve a discretized 
version of the problem with linear time complexity, coupled with 
iterative refinement at higher and higher resolutions.
Our discretized formulation can be thought of as generalizing 
the Itakura parallelogram \cite{itakura1975minimum};
the iterated refinement scheme is similar in nature to FastDTW
\cite{salvador2007toward}.
We offer our implementation as open source C++ code with an intuitive 
Python package called \gdtw{} that runs 50x faster
than other methods on standard problem sizes. 

We describe several extensions and variations of our method.
In one extension, we extend our optimization framework to find 
a time-warped center of or template for a set of signals; 
in a further extension, we cluster a set of signals into groups,
each of which is time-warped into one of a set of templates or prototypes.

\section{Dynamic time warping}\label{s-dtw}

\paragraph{Signals.}
A (vector-valued) signal $f$ is a function $f:[a,b] \to \reals^d$,
with argument time.
A signal can be specified or described in many ways,
for example a formula, or via a sequence of samples along with a method
for interpolating the signal values in between samples.
For example we can describe a signal as taking values 
$s_1, \ldots, s_N \in \reals^d$, 
at points (times)
$a\leq t_1 < t_2 < \cdots < t_N \leq b$, 
with linear interpolation in between these values and a constant extension
outside the first and last values:
\[
	f(t)= \left\{ \begin{array}{ll} s_1 & a \leq t < t_1\\
\frac{t_{i+1}-t}{t_{i+1}-t_i} s_{i} + 
\frac{t-t_i}{t_{i+1}-t_i} s_{i+1} & t_i \leq t < t_{i+1}, \quad i=1, \ldots,
N-1,\\
	                                s_N & t_N < t \leq b,
\end{array}\right.
\]

For simplicity, we will consider signals on the time 
interval $[0,1]$.

\paragraph{Time warp function.}
Suppose $\phi:[0,1]\to [0,1]$ is increasing, with $\phi(0)=0$ and
$\phi(1)=1$. We refer to $\phi$ as the \emph{time warp function},
and $\tau = \phi(t)$ as the warped time associated with real or 
original time $t$.
When $\phi(t)=t$ for all $t$, the warped time is the same as the original
time.  In general we can think of 
\[
	\tau-t = \phi(t) -t
\]
as the amount of cumulative warping at time $t$, and 
\[
	\frac{d\tau}{d t}(\tau-t)= \phi'(t)-1
\]
as the instantaneous rate of time warping at time $t$.
These are both zero when $\phi(t)=t$ for all $t$.

\paragraph{Time-warped signal.}
If $x$ is a signal, we refer to the signal $\tilde x = x\circ \phi$, \ie,
\[
	\tilde x(t) = x(\tau)=x(\phi(t)),
\]
as the \emph{time-warped} signal, or the time-warped version of 
the signal $x$.

\paragraph{Dynamic time warping.}
Suppose we are given two signals $x$ and $y$.
Roughly speaking, the dynamic time warping problem is to find a
warping function $\phi$ so that $\tilde x = x\circ \phi \approx y$.
In other words, we wish to warp time so that the time-warped version 
of the first signal is close to the second one.
We refer to the signal $y$ as the \emph{target}, since the goal is
warp $x$ to match, or align with, the target.



\begin{figure}[!ht]
\begin{subfigure}{.5\textwidth}
\begin{center}
\includegraphics[width=.9310\linewidth,trim={60 10 0 15}]{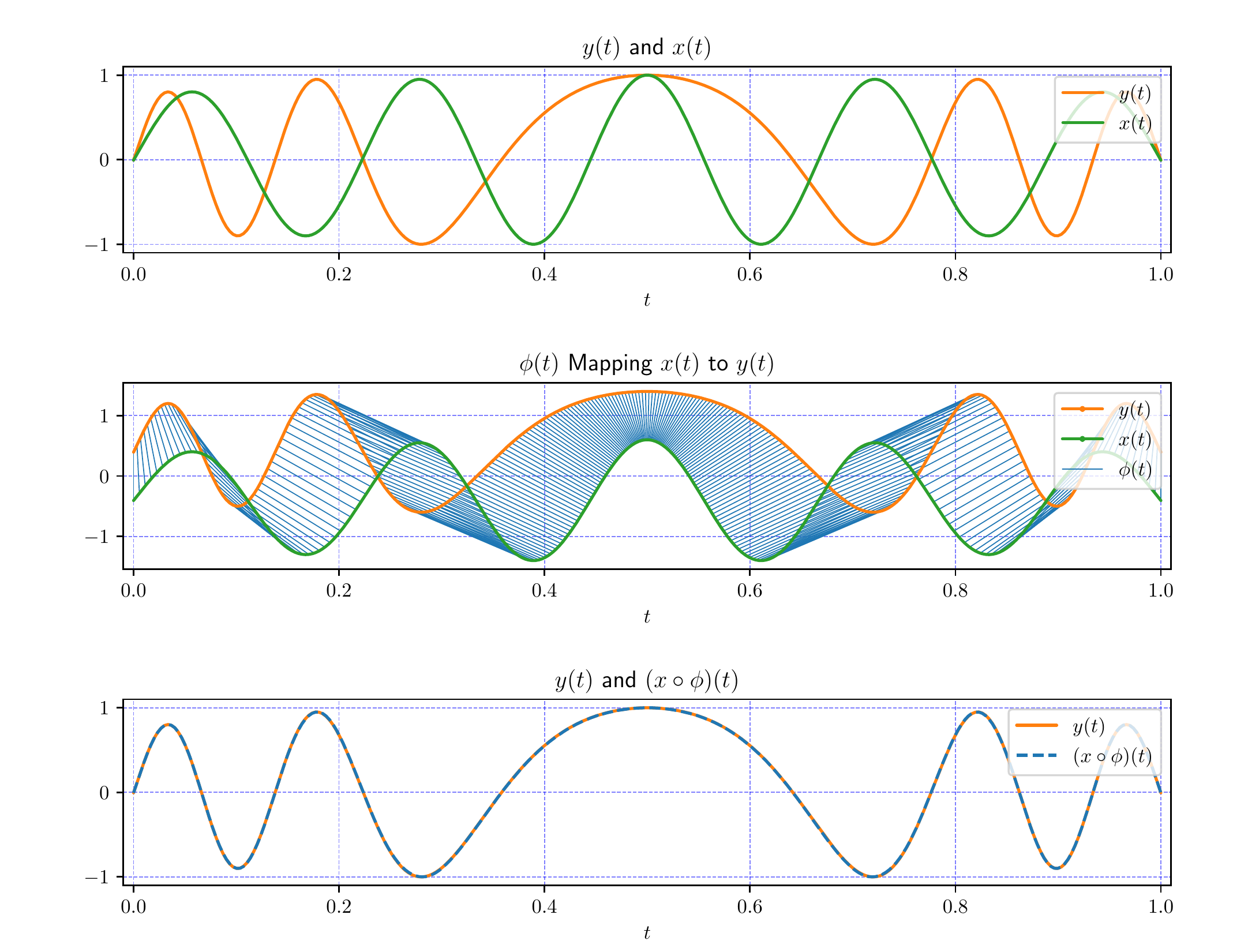}
\end{center}
\captionof{figure}{\emph{Top.} $x$ and $y$. \emph{Middle.} $\phi$. \emph{Bottom.} $\tilde x$ and $y$.}
\label{fig:example_functions}
\end{subfigure}
\begin{subfigure}{.5\textwidth}
\begin{center}
\includegraphics[width=1.03\linewidth,trim={30 50 30 16},clip]{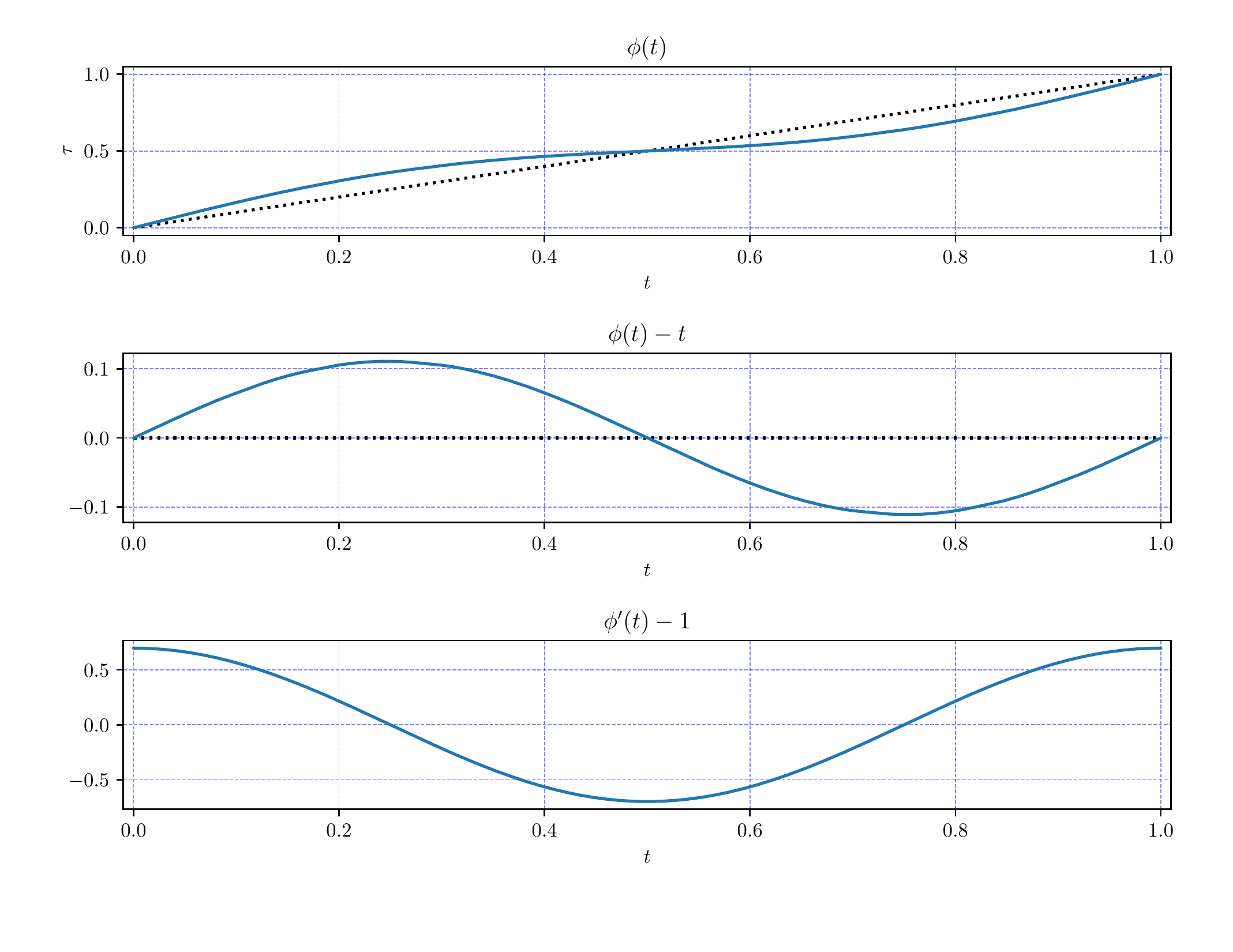}
\end{center}
\captionof{figure}{\emph{Top.} $\phi(t)$. \emph{Middle.} $\phi(t)-t$. \emph{Bottom.} $\phi'(t) - 1$.}
\label{fig:example_warps}
\end{subfigure}
\end{figure}

\paragraph{Example.}
An example is shown in figure~\ref{fig:example_functions}.
The top plot shows a scalar signal $x$ and target signal $y$, and the 
bottom plot shows the time-warped signal $\tilde x = x\circ \phi$ and $y$.
The middle plot shows the correspondence between $x$ and $y$ 
associated with the warping function $\phi$.
Figure~\ref{fig:example_warps} shows the time warping function; the 
next plot is the cumulative warp, and the next is the instantaneous 
rate of time warping.

\section{Optimization formulation}\label{s-formulation}

We will formulate the dynamic time warping problem as an 
optimization problem, where the time warp function $\phi$ is
the (infinite-dimensional) optimization variable to be chosen.
Our formulation is very similar to those used in machine learning,
where a fitting function is chosen to minimize an objective that 
includes a \emph{loss function} that measures the error in fitting the 
given data, and \emph{regularization terms} that penalize the 
complexity of the fitting function \cite{friedman2001elements}.

\paragraph{Loss functional.}
Let $L: \reals^d \to \reals$
be a \emph{vector penalty function}.
We define the \emph{loss} associated with a time warp function $\phi$,
on the two signals $x$ and $y$, as
\BEQ
\mathcal{L}(\phi) = \int_{0}^{1} L(x(\phi(t))- y(t)) \dt,
\label{eq:loss_cont}
\EEQ
the average value of the penalty function of the difference between
the time-warped first signal and the second signal.
The smaller $\mathcal L(\phi)$ is, the better we consider
$\tilde x =x\circ \phi$ to approximate $y$.

Simple choices of the penalty include $L(u)=\|u\|_2^2$ or 
$L(u)=\|u\|_1$.  The corresponding losses are the mean-square 
deviation and mean-absolute deviation, respectively.
One useful variation is the Huber penalty 
\cite{huber2011robust,boyd2004convex},
\[
	L(u)= \left\{ \begin{array}{ll} \|u\|_2^2 & \|u\|_2 \leq M\\
	                           2M \|u\|_2-M^2 & \|u\|_2 > M,
\end{array}\right.
\]
where $M>0$ is a parameter.  The Huber penalty coincides with the
least squares penalty for small $u$, but grows more slowly 
for $u$ large, and so is less sensitive to outliers.
Many other choices are possible, for example
\[
	L(u)= \left\{ \begin{array}{ll} 0 & \|u\| \leq \epsilon\\
	                                1 & \mbox{otherwise},
\end{array}\right.
\]
where $\epsilon$ is a positive parameter.
The associated loss $\mathcal L(\phi)$ is the fraction of time
the time-warped signal is farther than $\epsilon$ from the second
signal (measured by the norm $\| \cdot \|$).

The choice of penalty function $L$ 
(and therefore loss functional $\mathcal L$) will influence the
warping found, and should be chosen to capture the notion of
approximation appropriate for the given application.

\paragraph{Cumulative warp regularization functional.}
We express our desired qualities for or requirements on the time warp 
function using a regularization functional for the cumulative warp,
\BEQ
\calRcum(\phi) = \int_{0}^{1} \Rcum(\phi(t)-t) \dt,
\label{eq:reg_cum}
\EEQ
where $\Rcum: \reals \to \reals \cup \{ \infty \}$
is a penalty function on the cumulative warp.
The function $\Rcum$ can take on the value $+\infty$,
which allows us to encode constraints on $\phi$.
While we do not require it, we typically have $\Rcum(0)=0$, 
\ie, there is no cumulative regularization cost when 
the warped time and true time are the same.

\paragraph{Instantaneous warp regularization functional.}
The regularization functional for the instantaneous warp is
\BEQ
\calRinst(\phi) = \int_{0}^{1} \Rinst(\phi'(t)-1) \dt,
\label{eq:reg_inst}
\EEQ
where $\Rinst: \reals \to \reals \cup \{ \infty \}$ is the
penalty function on the instantaneous rate of 
time warping.
Like the function $\Rcum$, $\Rinst$ can take on the value $+\infty$,
which allows us to encode constraints on $\phi'$.
By assigning $\Rinst(u)=+\infty$ for $u<\Smin$, for example, 
we require that $\phi'(t)\geq \Smin$ for all $t$.
We will assume that this is the case for some positive $\Smin$,
which ensures that $\phi$ is invertible.
While we do not require it, we typically have $\Rinst(0)=0$,
\ie, there is no instantaneous regularization cost when the instantaneous 
rate of time warping is one.

As a simple example, we might choose
\[
\Rcum(u) = u^2, \qquad \Rinst(u) =
\left\{ 
\begin{array}{ll} 
	   u^2 & \Smin \leq u \leq \Smax\\
	\infty & \mbox{otherwise},
\end{array}\right.
\]
\ie, a quadratic penalty on cumulative warping, and a square penalty
on instantaneous warping, plus the constraint that the 
slope of $\phi$ must be between $\Smin$ and $\Smax$.
A very wide variety of penalties can be used to express our 
wishes and requirements on the warping function.

\paragraph{Dynamic time warping via regularized loss minimization.}
We propose to choose $\phi$ by solving the optimization problem
\BEQ
\begin{array}{ll} 
\mbox{minimize} &
f(\phi) = \mathcal{L}(\phi) + \Lamcum \calRcum (\phi) + 
\Laminst\calRinst (\phi) \\
\mbox{subject to} & \phi(0)=0, \quad \phi(1)=1,
\end{array}
\label{eq:obj_cont}
\EEQ
where $\Lamcum$ and $\Laminst$ are positive hyper-parameters used to
vary the relative weight of the three terms.
The variable in this optimization problem is the time warp function $\phi$.

\paragraph{Optimal control formulation.}
The problem~(\ref{eq:obj_cont})
is an infinite-dimensional, and generally non-convex,
optimization problem. 
Such problems are generally impractical to solve exactly,
but we will see that this particular problem can be efficiently 
and practically solved.

It can be formulated as a classical continuous-time optimal 
control problem \cite{bertsekas2005dynamic},
with scalar state $\phi(t)$ and action or input $u(t) = \phi'(t)$:
\BEQ\label{eq:opt_contr}
\begin{array}{ll} \mbox{minimize} & 
\int_0^1 \left( \ell(\phi(t),u(t),t) + 
\Laminst \Rinst(u(t))\right) \dt \\
\mbox{subject to} & \phi(0)=0, \quad \phi(1)=1, \quad \phi'(t)=u(t),
\quad 0\leq t \leq 1,
\end{array}
\EEQ
where $\ell$ is the state-action cost function
\[
	\ell(u,v,t) = L(x(u)- y(t)) + \Lamcum \Rcum(u).
\]

There are many classical methods for numerically
solving the optimal control problem~(\ref{eq:opt_contr}),
but these generally make strong assumptions
about the loss and regularization functionals (such as
smoothness), and do not solve the problem globally. 
We will instead solve~(\ref{eq:opt_contr}) by brute force dynamic 
programming, which is 
practical since the state has dimension one, and so can be discretized.

\paragraph{Lasso and ridge regularization.}
Before describing how we solve the optimal control 
problem~(\ref{eq:opt_contr}),
we mention two types of regularization that are widely used in machine
learning, and what types of warping functions typically result 
when using them.
They correspond to $\Rcum$ and $\Rinst$ being either $u^2$ 
(quadratic, ridge, or Tikhonov regularization \cite{TA77,hansen2005rank})
or $|u|$ (absolute value, $\ell_1$ regularization, or Lasso
\cite[p564]{golub2012matrix} \cite{tibshirani1996regression})

With $\Rcum(u)=u^2$, the regularization discourages large deviations
between $\tau$ and $t$, 
but the not the rate at which $\tau$ changes with $t$.
With $\Rinst(u)=u^2$, the regularization discourages large instantaneous 
warping rates.  
The larger $\Lamcum$ is, the less $\tau$ deviates from $t$;
the larger $\Laminst$ is, the more smooth the time warping function
$\phi$ is.

Using absolute value regularization is more interesting.  
It is well known in machine learning that using absolute value or 
$\ell_1$ regularization leads to solutions with an argument of the 
absolute value that is sparse, that is, often zero \cite{boyd2004convex}.
When $\Rcum$ is the absolute value, we can expect many times when
$\tau = t$, that is, the warped time and true time are the same.
When $\Rinst$ is the absolute value, we can expect many times when
$\phi'(t) =1$, that is, the instantaneous rate of time warping is zero.
Typically these regions grow larger as we increase the hyper-parameters
$\Lamcum$ and $\Laminst$.

\paragraph{Discretized time formulation.}

To solve the problem~(\ref{eq:obj_cont})
we discretize time with the $N$ values
\[
	0 = t_1 < t_2 < \cdots < t_N = 1.
\]
We will assume that $\phi$ is piecewise linear with knot points
at $t_1, \ldots, t_N$; to describe it we only need to
specify the warp values $\tau_i = \phi(t_i)$ for $i=1, \ldots, N$,
which we express as a vector $\tau \in \reals^N$.
We assume that the points $t_i$ are closely enough spaced that
the restriction to piecewise linear form is acceptable.
The values $t_i$ could be taken as the values at which the signal $y$
is sampled (if it is given by samples), or just the default linear 
spacing, $t_i = (i-1)/(N-1)$.
The constraints $\phi(0)=0$ and $\phi(1)=1$ are expressed as $\tau_1=0$
and $\tau_N=1$.

Using a simple Riemann approximation of the integrals and
the approximation 
\[
	\phi'(t_i) = \frac{\phi(t_{i+1})-\phi(t_i)}{t_{i+1}-t_i} =
	\frac{\tau_{i+1}-\tau_i}{t_{i+1}-t_i}, \quad 
	i=1, \ldots, N-1,
\]
we obtain the discretized objective
\BEQ \label{eq:discr_obj}
\hat f (\tau)  =
\sum_{i=1}^{N-1} (t_{i+1} - t_i) \left(
L(x(\tau_i)-y(t_i))
+ \Lamcum \Rcum(\tau_i - t_i)
+ \Laminst \Rinst \left( 
\frac{\tau_{i+1} - \tau_{i}}{t_{i+1} - t_i}\right) \right).
\EEQ
The discretized problem is to choose the vector $\tau \in \reals^N$
that minimizes $\hat f(\tau)$, subject to $\tau_1=0$, $\tau_N=1$.
We call this vector $\tau^\star$, with which we can construct
an approximation to function $\phi$ using piecewise-linear interpolation.
The only approximation here is the discretization; we 
can use standard techniques based on bounds on derivatives of the 
functions involved to bound the deviation between the continuous-time 
objective $f(\phi)$ and its discretized approximation $\hat f(\tau)$.

\section{Dynamic programming with refinement}\label{s-dp}

In this section we describe a simple method to 
minimize $\hat f(\tau)$ subject to $\tau_1 =0$ and $\tau_N=1$, \ie,
to solve the optimal control problem~(\ref{eq:opt_contr}) to obtain $\tau^*$.
We first discretize the possible values of $\tau_i$, whereupon the
problem can be expressed as a shortest path problem on a graph, and then 
efficiently and globally solved using standard dynamic 
programming techniques. To reduce the error associated with the
discretization of the values of $\tau_i$, we choose a new discretization
with the same number of values, but in a reduced range (and therefore,
more finely spaced values) around the previously
found values. This refinement converges in a few steps to a highly 
accurate solution of the discretized problem.
Subject only to the reasonable assumption that the discretization 
of the original time and warped time are sufficiently fine, 
this method finds the global solution.

\subsection{Dynamic programming}\label{s-dp-with-refinement}

We now discretize the values that $\tau_i$ is allowed to take:
\[
	\tau_i \in \mathcal T_i = \{\tau_{i1}, \ldots, \tau_{iM}\},
	\quad i=1,\ldots, N.
\]
One choice for these discretized values is linear spacing between
given lower and upper bounds on $\tau_i$, $0\leq l_i \leq u_i\leq 1$:
\[
	\tau_{ij} = l_i + \frac{j-1}{M-1}
	(u_i-l_i), \quad j=1,\ldots, M, \quad i=1,\ldots, N.
\]
Here $M$ is the number of values that we use to discretize each value of 
$\tau_i$ (which we take to be the same for each $i$, for simplicity).
We will assume that $0 \in \mathcal T_1$ and $1\in \mathcal T_N$,
so the constraints $\tau_1=0$ and $\tau_N=1$ are feasible.

\begin{figure}[!ht]
\begin{center}
\includegraphics[width=\linewidth]{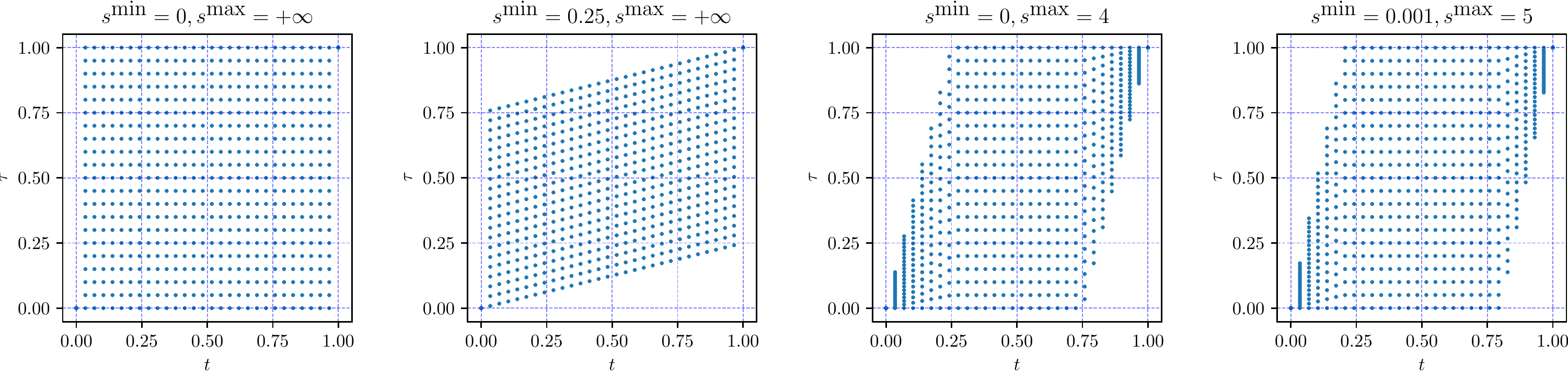}
\end{center}
\caption{  
	\emph{Left}. Unconstrained grid.
	\emph{Left center}. Effect of introducing $\Smin$. 
	\emph{Right center}. Effect of $\Smax$. 
	\emph{Right}. Typical parameters that work well for our method. 
}
\label{fig:taus}
\end{figure}

The bounds can be chosen as
\BEQ \label{eq:tau_bounds}
	l_i = \max\{ \Smin t_i , 1-\Smax (1-t_i) \}, 
	\quad
	u_i = \min\{ \Smax t_i , 1-\Smin (1-t_i) \}, 
	\quad i=1,\ldots, N,
\EEQ
where $\Smin$ and $\Smax$ are the given 
minimum and maximum allowed values of $\phi'$.
This is illustrated in figure~\ref{fig:taus}, where the nodes of
$\mathcal T$ are drawn at position $(t_i,\tau_{ij})$, for $N=30, M=20$ 
and various values of $\Smin$ and $\Smax$.
Note that since $\frac{N}{M}$ is the minimum slope, 
$M$ should be chosen to satisfy $M < \frac{N}{\Smax}$, a consideration
that is automated in the provided software.

The objective~(\ref{eq:discr_obj}) splits into a sum of 
terms that are functions of $\tau_i$, and terms that are functions
of $\tau_{i+1}-\tau_i$.
(These correspond to the separable state-action loss function terms
in the optimal control problem associated with $\phi(t)$ and
$\phi'(t)$, respectively.)
The problem is then globally solved by standard methods of dynamic
programming \cite{bellman2015applied}, using the methods we now describe.

We form a graph with $MN$ nodes, associated with the 
values $\tau_{ij}$, $i=1,\ldots, N$ and $j=1,\ldots, M$.
(Note that $i$ indexes the discretized values of $t$, and
$j$ indexes the discretized values of $\tau$.)
Each node $\tau_{ij}$ with $i<N$ has $M$ outgoing edges that terminate
at the nodes of the form $\tau_{i+1,k}$ for $k=1, \ldots, M$.
The total number of edges is therefore $(N-1)M^2$.
This is illustrated in figure~\ref{fig:taus} for $M=25$ and $N=100$,
where the nodes are shown at the location $(t_i,\tau_{ij})$.
(In practice $M$ and $N$ would be considerably larger.)

At each node $\tau_{ij}$ we associate the node cost 
\[
	L(x(\tau_{ij})-y(t_i)) + \Lamcum \Rcum(\tau_{ij}) 
\]
and on the edge from $\tau_{ij}$ to $\tau_{i+1,k}$ 
we associate the edge cost
\[
	\Laminst \Rinst 
	\left( 
		\frac{ \tau_{i+1,k} - \tau_{ij} }
			 {    t_{i+1}   -    t_{i}} 
	\right).
\]
With these node and edge costs, the objective $\hat f(\tau)$ is 
the total cost of a path starting at node $\tau_{11}=0$ and ending at 
$\tau_{NM}=1$.  
(Infeasible paths, for examples ones for which $\tau_{i+1,k}<\tau_{i,j}$,
have cost $+\infty$.)
Our problem is therefore to find the shortest
weighted path through a graph, which is readily done by 
dynamic programming.

The computational cost of dynamic programming is order $NM^2$ flops 
(not counting the evaluation of the loss and regularization terms).
With current hardware, it is entirely practical for $M=N=1000$ or 
even (much) larger.
The path found is the globally optimal one, \ie, $\tau^*$ 
minimizes $\hat f(\tau)$, subject to the discretization
constraints on the values of $\tau_i$.

\subsection{Iterative refinement}\label{s-iterative-refinement}

After solving the problem above by dynamic programming, we can
reduce the error induced by discretizing the values of $\tau_i$
by updating $l_i$ and $u_i$.
We shrink them both toward the current value of $\tau^*_i$, thereby 
reducing the gap between adjacent discretized values and reducing
the discretization error.
One simple method for updating the bounds is to reduce the range
$u_i-l_i$ by a fixed fraction $\eta$, say $1/2$ or $1/8$.

\begin{figure}[!ht]
\begin{center}
\includegraphics[width=\linewidth,trim={0 0 0 15},clip]{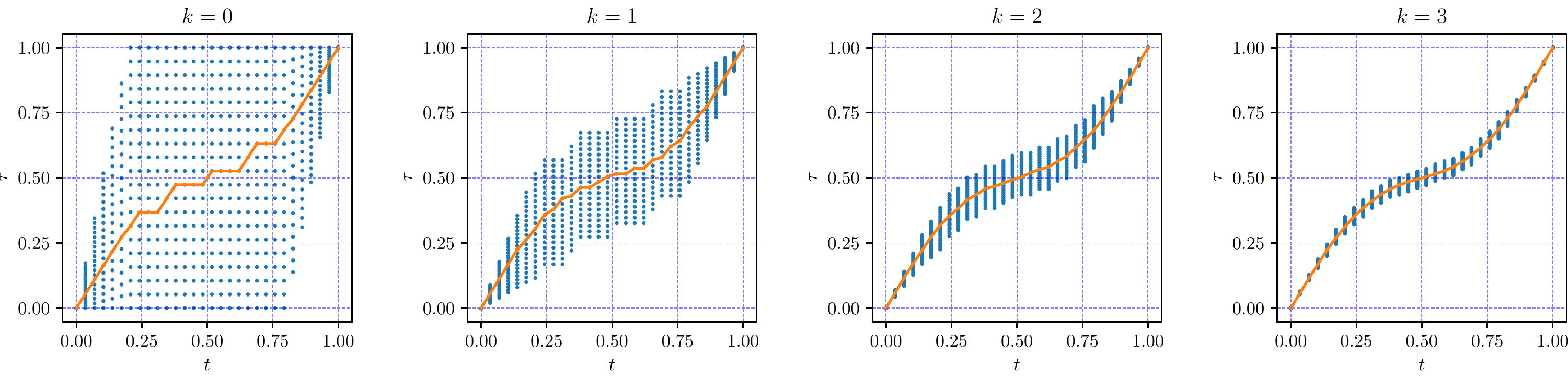}
\end{center}
\caption{ 
	\emph{Left to right.} Iterative refinement of $\tau^*$ for iterations
	$q=0,1,2,3$, with $\tau^*$ colored orange.
}
\label{fig:example_iterative_refinement}
\end{figure}

To do this we set
\[
	l_i^{(q+1)} = \max\{ 
		\tau^{*(q)}_i - \eta \frac{u_i^{(q)} - l_i^{(q)}}{2}, 
		l_i^{(0)} 
	\}, 
	\qquad
	u_i^{(q+1)} = \min\{ 
		\tau^{*(q)}_i + \eta \frac{u_i^{(q)} - l_i^{(q)}}{2}, 
		u_i^{(0)}
	\}
\]
in iteration $q+1$, where the superscripts in parentheses above 
indicate the iteration.
Using the same data as figure~\ref{fig:example_warps}, 
figure~\ref{fig:example_iterative_refinement} shows the iterative
refinement of $\tau^*$.
Here, nodes of $\mathcal T$ are plotted at position $(t_i,\tau_{ij})$,
as it is iteratively refined around $\tau^*_i$.

\subsection{Implementation}\label{s-implementation}

\paragraph{GDTW package.}
The algorithm described above has been implemented as the open source
Python package \gdtw{}, with the dynamic programming portion 
written in C++ for improved efficiency.
The node costs are computed and stored in an $M\times N$ array, and
the edge costs are computed on the fly and stored in an 
$M\times M\times N$ array.
For multiple iterations on group-level alignments (see \S\ref{s-alignment}),
multi-threading is used to distribute the program onto worker threads.

\paragraph{Performance.}
We give an example of the performance attained by \gdtw{} using
real-world signals described in \S\ref{s-examples}, which are 
uniformly sampled with $N=1000$.
Although it has no effect on method performance, we take
square loss, square cumulative warp regularization, and
square instantaneous warp regularization.
We take $M=100$. 

The computations are carried on a 4 core MacBook.
To compute the node costs requires 0.0055 seconds,
and to compute the shortest path requires 0.0832 seconds.
With refinement factor $\eta = .15$, only three iterations are needed
before no significant improvement is obtained,
and the result is essentially the same with other choices for the 
algorithm parameters $N$, $M$, and $\eta$.
Over 10 trials, our method only took an average of 0.25 seconds,
a 50x speedup over \texttt{FastDTW}, which took an average of 14.1 seconds 
to compute using a radius of 50, which is equivalent to $M=100$. 
All of the data and example code necessary to reproduce these results 
are available in the \gdtw{} repository.
Also available are supplementary materials that contain step-by-step 
instructions and demonstrations on how to reproduce these results.

\subsection{Validation}

To test the generalization ability of a specific time warping model,
parameterized by $L,\Lamcum,\Rcum,\Laminst,$ and $\Rinst$,
we use out-of-sample validation
by randomly partitioning $N$ discretized time values 
$0=t_1, \ldots, t_N=1$
into two sorted ordered sets that contain the boundaries, $\Ttrain \cup \{ 0, 1 \}$ and 
$\Ttest \cup \{ 0, 1 \}$.
Using only the time points in $\Ttrain$, we obtain our time warping 
function $\phi$ by minimizing our discretized 
objective~(\ref{eq:discr_obj}).
(Recall that our method does not require signals to be sampled at 
regular intervals, and so will work with the irregularly spaced time 
points in $\Ttrain$.)

\begin{figure}[!ht]
\begin{center}
\includegraphics[width=.5\linewidth,trim={0 0 0 0},clip]{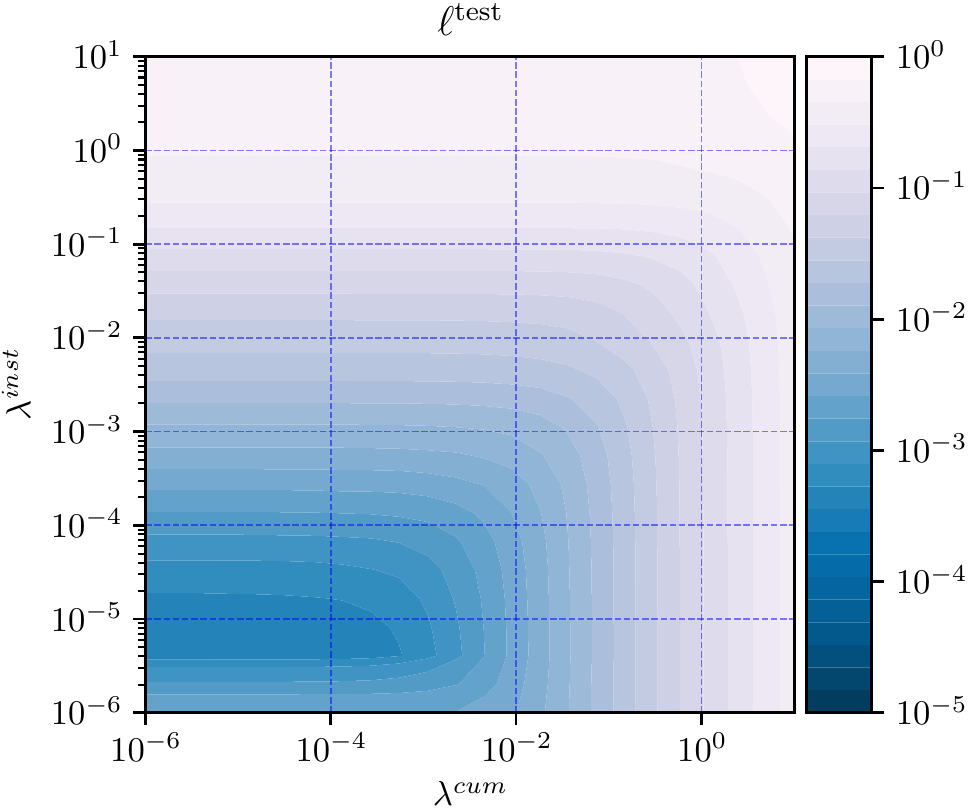}\vspace{1.5mm}
\end{center}
\caption{
	Test loss.
}
\label{fig:example_validation_grid}
\end{figure}

We compute two loss values: a training error  
\[
	\ell^{\mathrm{train}} = 
	\sum_{i=1}^{|\Ttrain|-1} (\Ttrain_{i+1} - \Ttrain_i) \left(
		L(x(\phi(\Ttrain_i)) - y(\Ttrain_i)) 
	\right),
\] 
and a test error 
\[
	\ell^{\mathrm{test}} = 
	\sum_{i=1}^{|\Ttest|-1} (\Ttest_{i+1} - \Ttest_i) \left(
		L(x(\phi(\Ttest_i)) - y(\Ttest_i)) 
	\right).
\]
Figure~\ref{fig:example_validation_grid} shows 
$\ell^{\mathrm{test}}$ over a grid of values of $\Lamcum$ and $\Laminst$,
for a partition where $\Ttrain$ and $\Ttest$ each contain $~50\%$ of the 
time points.
In this example, we use the signals shown figure~\ref{fig:example_functions}.

\begin{figure}[!ht]
\begin{center}
\includegraphics[width=.5\linewidth,trim={0 0 0 0},clip]{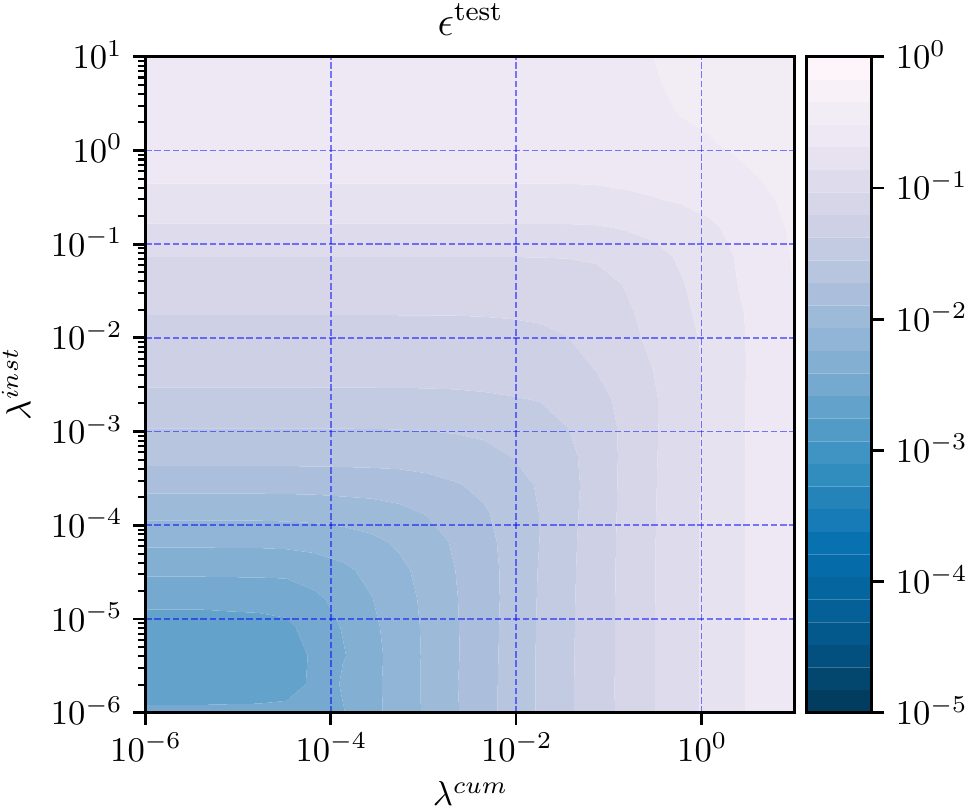}\vspace{1.5mm}
\end{center}
\caption{
	Test error.
}
\label{fig:example_error_grid}
\end{figure}

\paragraph{Ground truth estimation.}
When a ground truth warping function, $\Phitrue$, is available, 
we can score how well our $\phi$ approximates $\Phitrue$
by computing the following errors:
\[
	\epsilon^{\mathrm{train}} = 
	\sum_{i=1}^{|\Ttrain|-1} (\Ttrain_{i+1} - \Ttrain_i) \left(
		L(\Phitrue(\Ttrain_i) - \phi(\Ttrain_i)) 
	\right),
\]
and
\[
	\epsilon^{\mathrm{test}} = 
	\sum_{i=1}^{|\Ttest|-1} (\Ttest_{i+1} - \Ttest_i) \left(
		L(\Phitrue(\Ttest_i) - \phi(\Ttest_i)) 
	\right).
\]
In the example shown in 
figure~\ref{fig:example_validation_grid}, 
target signal $y$ is constructed by composing $x$ with a 
known warping function $\Phitrue$, such that $y(t) = (x \circ \Phitrue)(t)$.
Figure~\ref{fig:example_error_grid} shows the contours of 
$\epsilon^{\mathrm{test}}$ for this example.

\section{Examples}\label{s-examples}

\begin{figure}[!ht]
\begin{center}
\includegraphics[width=\linewidth,trim={0 30.25 0 0},clip]{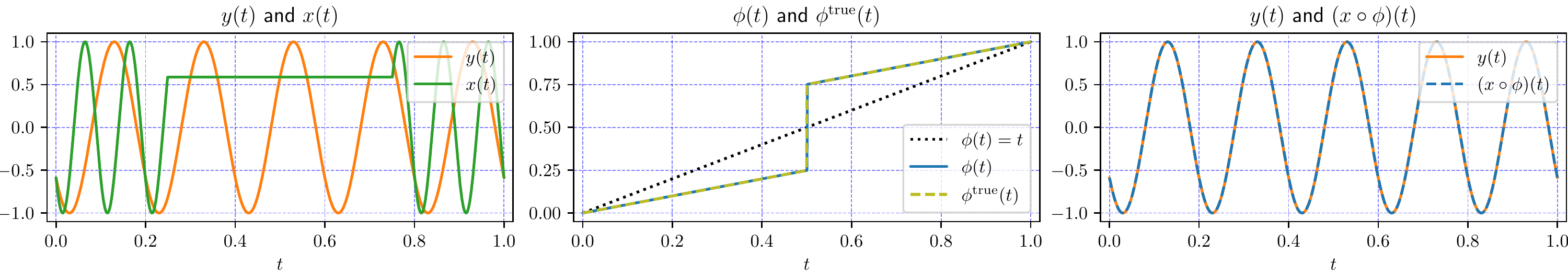}\vspace{1.5mm}
\includegraphics[width=\linewidth,trim={0 30.25 0 18},clip]{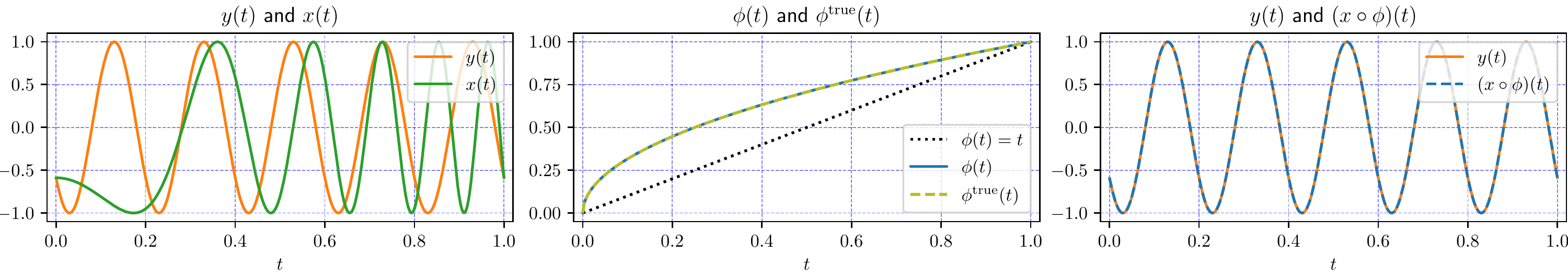}\vspace{1.5mm}
\includegraphics[width=\linewidth,trim={0 0 0 18},clip]{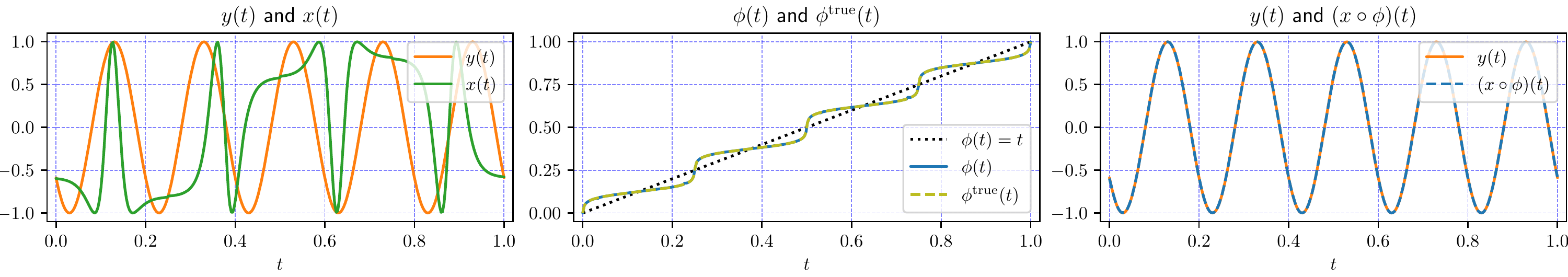}
\end{center}
\caption{
	\emph{Left.} Signal $x$ and target signal $y$. 
	\emph{Middle.} Warping function $\phi$ and the ground truth
	warping $\Phitrue$.
	\emph{Right.} The time-warped $x$ and $y$.\\
}
\label{fig:example_warping_functions}
\end{figure}

\begin{figure}[!ht]
\begin{center}
\includegraphics[width=\linewidth,trim={20 30.25 2 0},clip]{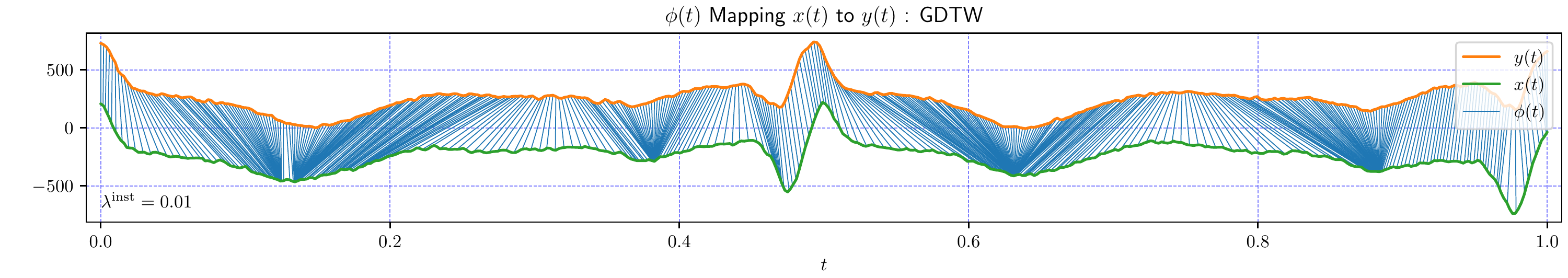}\vspace{1.5mm}
\includegraphics[width=\linewidth,trim={20 30.25 2 18},clip]{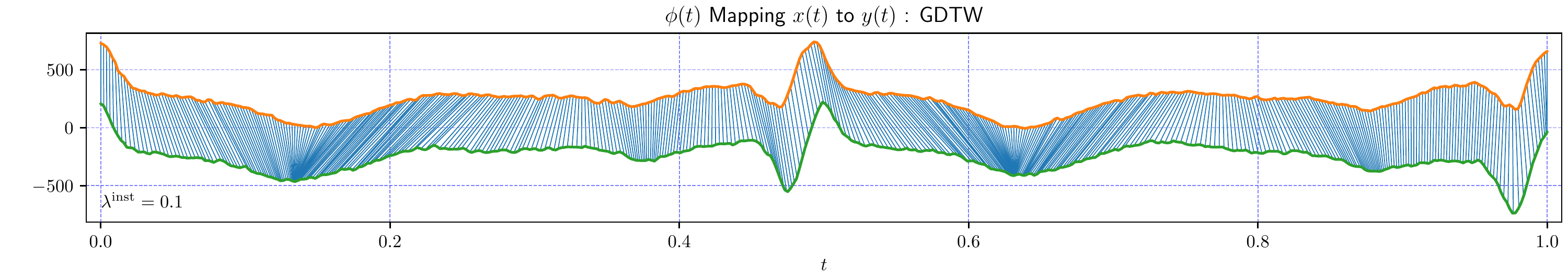}\vspace{1.5mm}
\includegraphics[width=\linewidth,trim={20 0 2 18},clip]{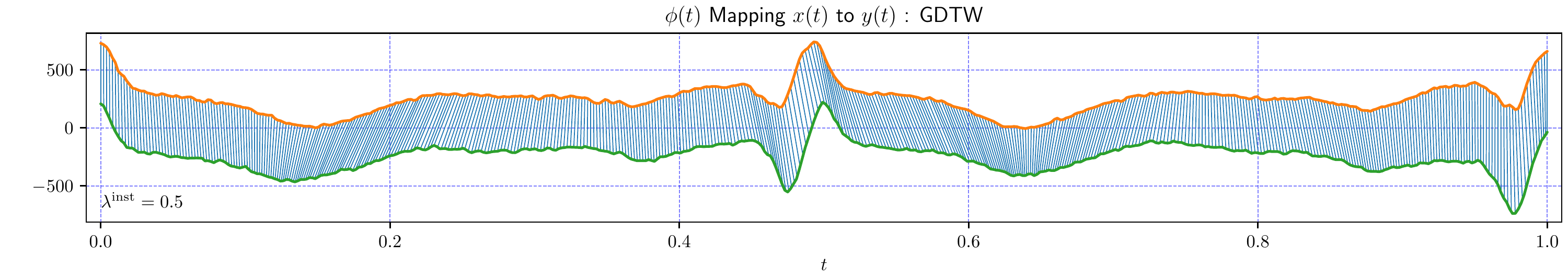}
\includegraphics[width=\linewidth,trim={20 240 2 240},clip]{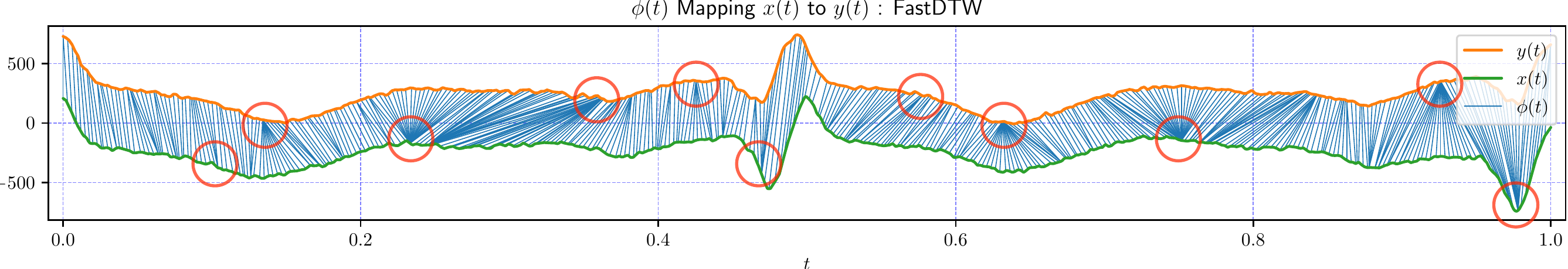}\vspace{1.5mm}
\end{center}
\caption{
	\emph{Top four}. ECGs warped using our method while increasing
	$\Laminst$.
	\emph{Bottom}. Results using FastDTW, with a few of the singularities circled in red.\\
}
\label{fig:example_ecg_lambda_ins}
\end{figure}

\begin{figure}[!ht]
\begin{center}
\includegraphics[width=\linewidth,trim={10 0 0 0},clip]{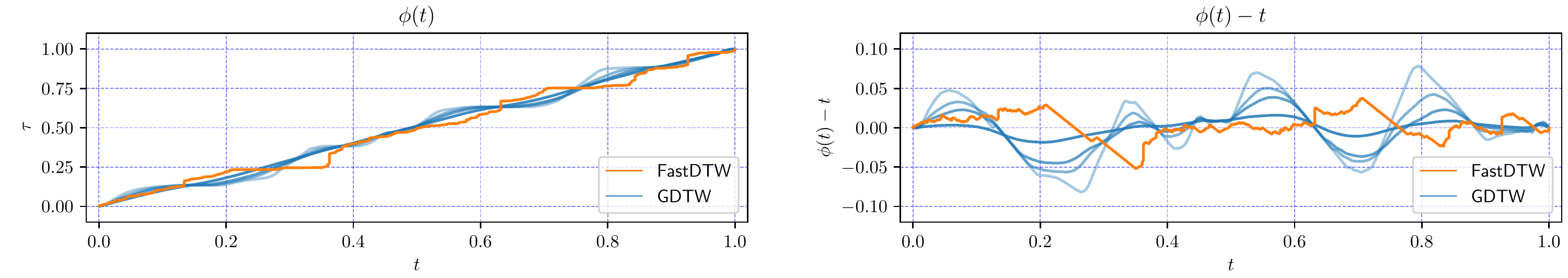}\vspace{1.5mm}
\end{center}
\caption{
	\emph{Left}. $\phi(t)$
	\emph{Right}. $\phi(t)-t$ for ECGs. 
	(Smoother lines correspond to larger $\Laminst$.)
}
\label{fig:example_ecg_lambda_ins_phis}
\end{figure}

We present a few examples of alignments using our method.
Figure~\ref{fig:example_warping_functions} is a synthetic example of 
different types of time warping functions. 
Figure~\ref{fig:example_ecg_lambda_ins} is real-world example 
using biological signals (ECGs).
We compare our method using varying amounts of regularization 
$\Laminst \in \{0.01,0.1,0.5 \},N=1000,M=100$ 
to those using with FastDTW~\cite{salvador2007toward}, 
as implemented in the Python package \texttt{FastDTW}~\cite{fastdtw}
using the equivalent graph size $N=1000,\mathrm{radius}=50$.
As expected, the alignments using regularization are smoother and less prone to 
singularities than those from FastDTW, which are unregularized.
Figure~\ref{fig:example_ecg_lambda_ins_phis} shows how the time warp 
functions become smoother as $\Laminst$ grows.

\section{Extensions and variations}

We will show how to extend our formulation to address complex scenarios, 
such as aligning a portion of a signal to the target,
regularization of higher-order derivatives, 
and symmetric time warping, where both signals align to each other.

\subsection{Alternate boundary and slope constraints}
\begin{figure}[!ht]
\begin{center}
\includegraphics[width=\linewidth]{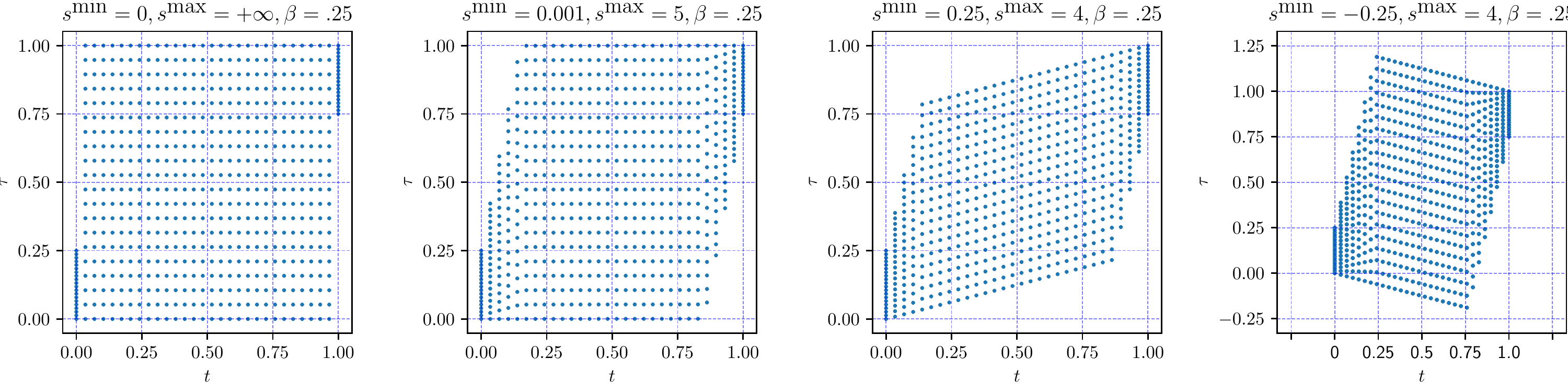}
\end{center}
\caption{  
	\emph{Left}. Effect of introducing $\beta$ to unconstrained grid.
	\emph{Left center}. Effect of introducing $\beta$ using typical parameters. 
	\emph{Right center}. Effect of introducing $\beta$ using larger $\Smin$. 
	\emph{Right}. Effect of negative $\Smin$. 
}
\label{fig:taus_extended}
\end{figure}

We can align a portion of a signal with the target by adjusting the 
boundary constraints to allow $0 \ge \phi(0) \ge \beta$ and 
$(1-\beta) \le \phi(1) \le 1$, 
for margin $\beta = \{ x \in \reals \ | \ 0 < x < 1 \}$.
We incorporate this by reformulating~(\ref{eq:tau_bounds}) as 
\[
	l_i = \max\{ \Smin t_i , (1-\Smax(1-t_i))-\beta \},
	\quad
	u_i = \min\{ \Smax t_i + \beta, 1 -\Smin (1-t_i) \}, 
	\quad i=1,\ldots, N.
\]
We can also allow the slope of $\phi$ to be negative, by choosing
$\Smin < 0$. 
These modifications are illustrated in figure~\ref{fig:taus_extended}, 
where the nodes of 
$\mathcal T$ are drawn at position $(t_i,\tau_{ij})$, for $N=30, M=20$ 
and various values of $\beta$, $\Smin$, and $\Smax$.

\subsection{Penalizing higher-order derivatives}
We can extend the formulation to include a constraint or objective term 
on the higher-order derivatives, such as the second derivative $\phi''$.
This requires us to extend the discretized state space to include not 
just the current $M$ values, but also the last $M$ values,
so the state space size grows to $M^2$ in the dynamic programming problem.

The regularization functional for the second-order instantaneous warp is
\[
\calRinstso(\phi) = \int_{0}^{1} \Rinstso(\phi''(t)) \dt,
\]
where $\Rinstso: \reals \to \reals \cup \{ \infty \}$ is the
penalty function on the second-order instantaneous rate of 
time warping.
Like the function $\Rinst$, $\Rinstso$ can take on the value $+\infty$,
which allows us to encode constraints on $\phi''$.

We use a three-point central difference approximation of the second 
derivative for evenly spaced time points
\[
	\phi''(t_i) = 
	\frac{\phi(t_{i+1})-2\phi(t_i)+\phi(t_{i-1})}{(t_{i+1}-t_{i})^2} =
	\frac{\tau_{i+1}-2\tau_i+\tau_{i-1}}{(t_{i+1}-t_{i})^2}, 
	\quad i=1, \ldots, N-1,
\]
and unevenly spaced time points
\[
	\phi''(t_i) = 
	\frac{2(\delta_1 \phi(t_{i+1}) - (\delta_1 + \delta_2)\phi(t_i) 
		+ \delta_2 \phi(t_{i-1}))}{\delta_1 \delta_2 (\delta_1 + \delta_2)} =
	\frac{2(\delta_1    \tau_{i+1} - (\delta_1 + \delta_2)\tau_i    
		+ \delta_2 \tau_{i-1})}{   \delta_1 \delta_2 (\delta_1 + \delta_2)},
\]
for $i=1, \ldots, N-1$, where $\delta_1 = t_{i}-t_{i-1}$ and 
$\delta_2 = t_{i+1}-t_{i}$.
With this approximation, we obtain the discretized objective
\[
\hat f (\tau)  =
\sum_{i=1}^{N-1} (t_{i+1} - t_i) \left(
L(x(\tau_i)-y(t_i))
+ \Lamcum \Rcum(\tau_i - t_i)
+ \Laminst \Rinst( \phi'(t_i) )
+ \Laminstso \Rinstso( \phi''(t_i) ) \right).
\]

\subsection{General loss}

The two signals need not be vector valued; they could have categorical
values, for example 
\[
L(\tau_i, t_i)= \left\{ \begin{array}{ll} 1 & \tau_i \neq t_i\\
                                          0 & \mbox{otherwise},
\end{array}\right.
\]
or 
\[
L(\tau_i, t_i)= \left\{ \begin{array}{ll} g(\tau_i,t_i) & \tau_i \neq t_i\\
                                          0 & \mbox{otherwise},
\end{array}\right.
\]
where $g:\reals_{++} \times \reals_{++} \to \reals$ is a 
\emph{categorical distance function} that can specify the cost of certain 
mismatches or a similarity matrix \cite{needleman1970general}. 

Another example could use the Earth mover's distance, 
$\mbox{EMD}:\reals^n \times \reals^n \to \reals$, 
between two short-time spectra
\[
L(\phi, t_i) = \mbox{EMD}( 
	\{ \phi(t_i - \rho), \ldots, \phi(t_i), \ldots, \phi(t_i + \rho) \},
	\{      t_i - \rho , \ldots,      t_i , \ldots,      t_i + \rho  \}
),
\]
where $\rho \in \reals$ is a radius around time point $t_i$.

\subsection{Symmetric time warping}
Until this point, we have used \emph{unidirectional} time warping, 
where signal $x$ is time-warped to align with $y$ such that 
$x\circ \phi \approx y$.
We can also perform \emph{bidirectional} time warping, 
where signals $x$ and $y$ are time-warped each other.
Bidirectional time warping results in two time warp functions, 
$\phi$ and $\psi$, where $x \circ \phi \approx y \circ \psi$.

Bidirectional time warping requires a different loss functional.
Here we define the \emph{bidirectional loss} associated with time warp 
functions $\phi$ and $\psi$, on the two signals $x$ and $y$, as
\[
\mathcal{L}(\phi,\psi) = \int_{0}^{1} L(x(\phi(t))- y(\psi(t))) \dt,
\] 
where we distinguish the bidirectional case by using two arguments, 
$\mathcal{L}(\phi,\psi)$, instead of one, $\mathcal{L}(\phi)$, as 
in~(\ref{eq:loss_cont}).

Bidirectional time warping can be \emph{symmetric} or \emph{asymmetric}.
In the symmetric case, we choose $\phi, \psi$ by solving the optimization 
problem
\[
\begin{array}{ll} 
\mbox{minimize} &
\mathcal{L}(\phi,\psi) + \Lamcum \calRcum (\phi) + 
\Laminst\calRinst (\phi) \\
\mbox{subject to} & \phi(0)=0, \quad \phi(1)=1, \quad \psi(t)=2t-\phi(t),
\end{array}
\]
where the constraint $\psi(t)=2t-\phi(t)$ ensures that $\phi$ and $\psi$ 
are symmetric
about the identity.
The symmetric case does not add additional computational complexity, 
and can be readily solved using the iterative refinement procedure 
described in~\S\ref{s-dp}.

In the asymmetric case, $\phi, \psi$ are chosen by solving the 
optimization problem
\[
\begin{array}{ll} 
\mbox{minimize} &
\mathcal{L}(\phi,\psi)
+ \Lamcum \calRcum (\phi) + \Lamcum \calRcum (\psi)
+ \Laminst\calRinst (\phi) + \Laminst\calRinst (\psi)\\
\mbox{subject to} & \phi(0)=0, \quad \phi(1)=1, \quad \psi(0)=0, \quad \psi(1)=1.
\end{array}
\]
The asymmetric case requires $\Rcum$, $\Rinst$ to allow negative slopes 
for $\psi$.
Further, it requires a modified iterative refinement procedure (not 
described here) with an increased complexity of order $NM^4$ flops, 
which is impractical when $M$ is not small.

\section{Time-warped distance, centering, and clustering}\label{s-alignment}

In this section we describe three simple 
extensions of our optimization formulation that yield useful
methods for analyzing a set of signals $x_1, \ldots, x_M$.

\subsection{Time-warped distance}\label{s-time-warped-distance}

For signals $x$ and $y$, we can interpret the optimal value 
of~(\ref{eq:obj_cont}) as the \emph{time-warped distance} between $x$ 
and $y$, denoted $D(x,y)$.
(Note that this distance measures takes into account both the loss and 
the regularization, which measures how much warping was needed.)
When $\Lamcum$ and $\Laminst$ are zero, we recover the unconstrained 
DTW distance~\cite{sakoe1978dynamic}.
This distance is not symmetric; we can (and usually do) have 
$D(x,y)\neq D(y,x)$.
If a symmetric distance is preferred, we can take 
$(D(x,y)+D(y,x))/2$, or the optimal value of the group alignment 
problem~(\ref{eq:gp_align}), with a set of original signals $x,y$.

The warp distance can be used in many places where a conventional 
distance between two signals is used.  
For example we can use warp distance to carry out $k$ nearest neighbors
regression \cite{xi2006fast} or classification.  
Warp distance can also be used to create features for further machine 
learning.  
For example, suppose that we have carried out clustering into $K$ 
groups, as discussed above, with target or group centers or exemplar
signals 
$y_1, \ldots, y_K$.
From these we can create a set of $K$ features related to the warp 
distance of a new signal $x$ to the centers $y_1, \ldots, y_K$, as
\[
	z_i = \frac{e^{d_i/\sigma}}{\sum_{j=1}^K e^{d_j/\sigma}}, 
	      \quad i=1, \ldots, K,
\]
where $d_i = D(x,y_i)$ and $\sigma$ is a positive (scale) 
hyper-parameter.

\subsection{Time-warped alignment and centering}\label{s-gp-align}

In \emph{time-warped alignment},
the goal is to find a common target signal $\mu$ that each of
the original signals can be warped to, at low cost.
We pose this in the natural way as the optimization problem
\BEQ
\begin{array}{ll} 
\mbox{minimize} &
\sum_{i=1}^M \left( \int_0^1 L(x_i(\phi_i(t)) - \mu(t)) \; dt + 
\Lamcum \calRcum (\phi_i) + 
\Laminst\calRinst (\phi_i) \right) \\
\mbox{subject to} & \phi_i(0)=0, \quad \phi_i(1)=1,
\end{array}
\label{eq:gp_align}
\EEQ
where the variables are the warp functions $\phi_1, \ldots, \phi_M$
and the target $\mu$, and
$\Lamcum$ and $\Laminst$ are positive hyper-parameters.
The objective is the sum of the objectives for time warping each $x_i$
to $\mu$.  
This is very much like our basic formulation~(\ref{eq:obj_cont}),
except that we have multiple signals to warp, and the target $\mu$ is 
also a variable that we can choose.

The problem~(\ref{eq:gp_align}) is hard to solve exactly, but a simple 
iterative procedure seems to work well.
We observe that if we fix the target $\mu$,
the problem splits into $M$ separate dynamic time warping problems
that we can solve (separately, in parallel) using the method described 
in \S\ref{s-dp}.
Conversely, if we fix the warping functions $\phi_1, \ldots, \phi_M$,
we can optimize over $\mu$ by minimizing
\[
	\sum_{i=1}^M \int_0^1 L(x_i(\phi_i(t)) - \mu(t)) \; dt.
\]
This is turn amounts to choosing each $\mu(t)$ to minimize
\[
	\sum_{i=1}^M L(x_i(\phi_i(t))- \mu(t)).
\]
This is typically easy to do; for example, 
with square loss,
we choose $\mu(t)$ to be the mean of $x_i(\phi_i(t))$;
with absolute value loss, 
we choose $\mu(t)$ to be the median of $x_i(\phi_i(t))$.

This method of alternating between updating the target $\mu$
and updating the warp functions (in parallel) typically converges 
quickly.
However, it need not converge to the global minimum.
One simple initialization is to start with no warping, \ie,
$\phi_i(t)=t$. 
Another is to choose one of the original signals as the initial value 
for $\mu$.

\begin{figure}[!ht]
\begin{center}
\includegraphics[width=\linewidth,trim={0 30.25 0 0},clip]{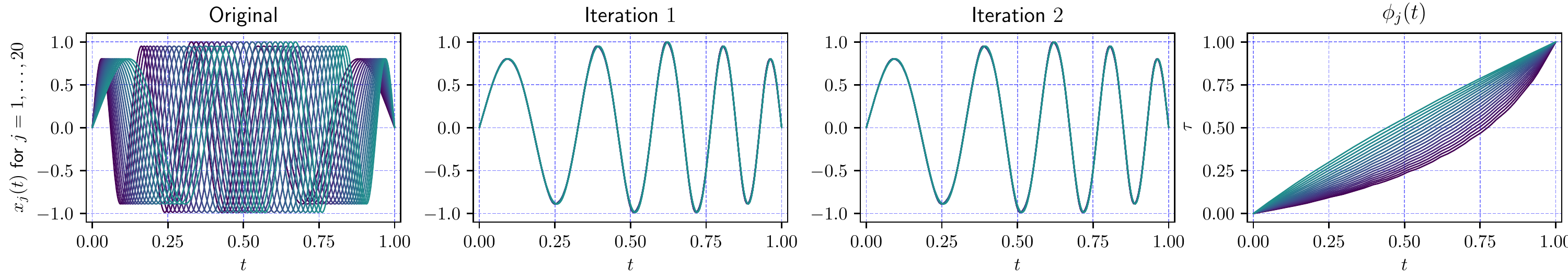}\vspace{1.5mm}
\includegraphics[width=\linewidth,trim={0 0 0 18},clip]{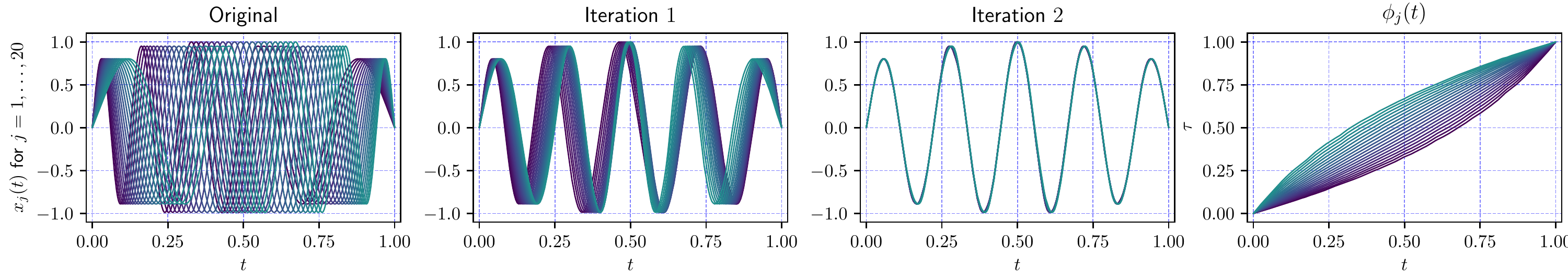}\vspace{1.5mm}
\end{center}
\caption{
	\emph{Top}. Time-warped mean.
	\emph{Bottom}. Centered time-warped mean.
	\emph{Left}. Original signals. 
	\emph{Left center}. Warped signals after iteration 1. 
	\emph{Right center}. Warped signals after iteration 2. 
	\emph{Right}. Time warp functions after iteration 2. 
}
\label{fig:centered_time_warped_mean}
\end{figure}

As a variation, we can also require the warping functions to be 
evenly arranged about a common time warp center, for example $\phi(t) = t$.
We can do this by imposing a ``centering" constraint on~(\ref{eq:gp_align}),
\BEQ
\begin{array}{ll} 
\mbox{minimize} &
\sum_{i=1}^M \left( \int_0^1 L(x_i(\phi_i(t)) - \mu(t)) \; dt + 
\Lamcum \calRcum (\phi_i) + 
\Laminst\calRinst (\phi_i) \right) \\
\mbox{subject to} & \phi_i(0)=0, \quad \phi_i(1)=1, 
	\quad \frac{1}{M}\sum_{i=1}^M \phi_i(t) = t ,
\end{array}
\label{eq:gp_centered_align}
\EEQ
where $\frac{1}{M}\sum_{i=1}^M \phi_i(t) = t$
forces $\phi_1, \ldots, \phi_M$ to be evenly 
distributed around the identity $\phi(t)=t$.
The resulting \emph{centered} time warp functions, can be used to produce 
a centered time-warped mean.
Figure~\ref{fig:centered_time_warped_mean} compares a time-warped mean
with and without centering, using synthetic data 
consisting of multi-modal signals from \cite{srivastava2011registration}.

\begin{figure}[!ht]
\begin{center}
\includegraphics[width=\linewidth,trim={0 30.25 0 0},clip]{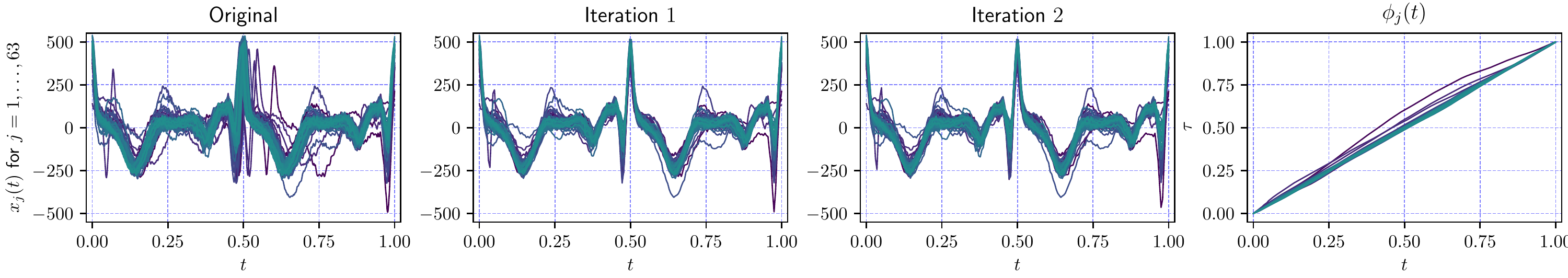}\vspace{1.5mm}
\includegraphics[width=\linewidth,trim={0 0 0 18},clip]{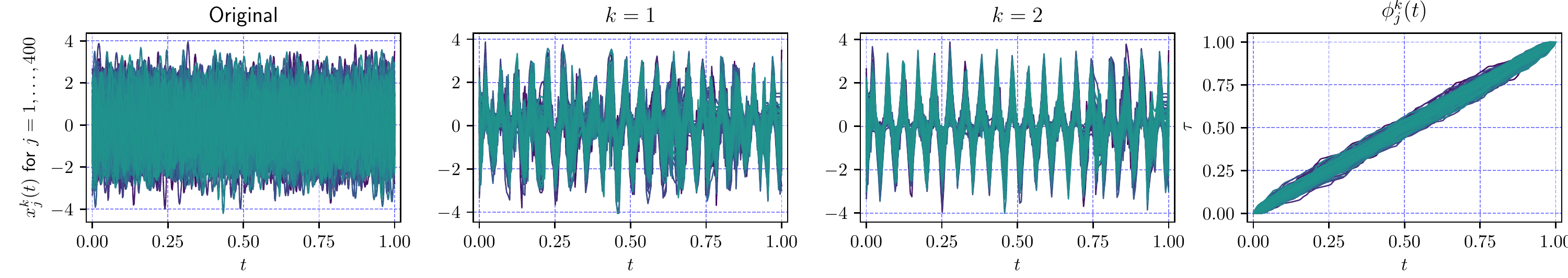}\vspace{1.5mm}
\end{center}
\caption{
	\emph{Top}. ECG signals.
	\emph{Bottom}. Engine sensor signals.
	\emph{Left}. Original signals. 
	\emph{Left center}. Warped signals after iteration 1. 
	\emph{Right center}. Warped signals after iteration 2. 
	\emph{Right}. Time warp functions after iteration 2. 
}
\label{fig:example_group_alignments}
\end{figure}

Figure~\ref{fig:example_group_alignments} shows examples of 
centered time-warped means of real-world data (using our default 
parameters), consisting of ECGs and sensor data from an automotive 
engine \cite{forddata}.
The ECG example demonstrates that subtle features of the input sequences
are preserved in the alignment process, and the engine example
demonstrates that the alignment process can find structure in noisy data.

\subsection{Time-warped clustering}\label{s-time-warped-clustering}

A further generalization of our optimization formulation allows us to
cluster set of signals $x_1, \ldots, x_M$ into $K$ groups, with each 
group having a template or center or exemplar.
This can be considered a time-warped version of $K$-means clustering;
see, \eg, \cite[Chapter 4]{boyd2018introduction}.
To describe the clusters we use the $M$-vector $c$, with $c_i = j$ 
meaning that signal $x_i$ is assigned to group $j$, 
where $j\in \{1,\ldots,M\}$.  
The exemplars or templates are the signals denoted $y_1, \ldots, y_K$.

\BEQ
\begin{array}{ll} 
\mbox{minimize} &
\sum_{i=1}^M \left( \int_0^1 L(x_i(\phi_i(t)) - y_{c_i}(t)) \; dt + 
\Lamcum \calRcum (\phi_i) + 
\Laminst\calRinst (\phi_i) \right) \\
\mbox{subject to} & \phi_i(0)=0, \quad \phi_i(1)=1,
\end{array}
\label{e-cluster}
\EEQ
where the variables are the warp functions $\phi_1, \ldots, \phi_M$,
the templates $y_1, \ldots, y_K$, and the assignment vector $c$.
As above, $\Lamcum$ and $\Laminst$ are positive hyper-parameters.

\begin{figure}[!ht]
\begin{center}
\includegraphics[width=\linewidth]{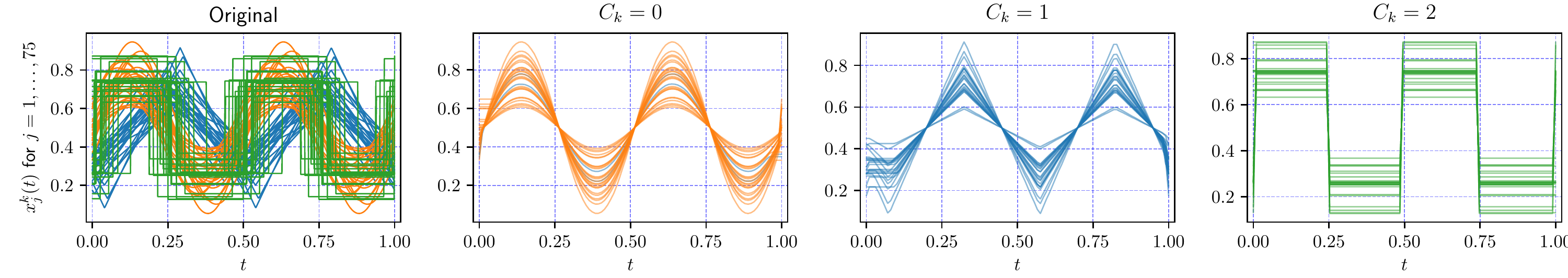}\vspace{1.5mm}
\end{center}
\caption{$K$-means alignment on synthetic data.}
\label{fig:example_kmeans}
\end{figure}

We solve this (approximately) by cyclically optimizing over 
the warp functions, the templates, and the assignments.
Figure~\ref{fig:example_kmeans} shows an example of this 
procedure (using our default parameters) 
on a set of sinusoidal, square, and triangular signals of 
varying phase and amplitude.

\section{Conclusion}\label{s-conclusion}

We claim three main contributions. 
We propose a full reformulation of DTW in continuous time that eliminates 
singularities without the need for preprocessing or step functions. 
Because our formulation allows for non-uniformly sampled signals, we 
are the first to demonstrate how validation can be used for DTW model selection.
Finally, we offer an implementation that runs 50x faster than 
state-of-the-art methods on typical problem sizes, and distribute our 
C++ code (as well as all of our example data) as an open-source Python 
package called \gdtw{}.

\clearpage
\bibliography{paper.bib}

\end{document}